\documentclass[letterpaper, 10 pt, conference]{ieeeconf}  

\IEEEoverridecommandlockouts                              

\usepackage{times}
\usepackage{epsfig}
\usepackage{graphicx}
\usepackage{amsmath}
\usepackage{amssymb}

\usepackage{bm}
\usepackage{makecell}
\usepackage{array}
\usepackage{booktabs,caption}
\newcolumntype{P}[1]{>{\centering\arraybackslash}p{#1}}
\newcolumntype{M}[1]{>{\centering\arraybackslash}m{#1}}
\usepackage{multirow}
\usepackage{subcaption}
\usepackage[normalem]{ulem}
\usepackage{soul} 
\usepackage{wrapfig}
\usepackage[pagebackref=true,breaklinks=true,letterpaper=true,colorlinks,bookmarks=false]{hyperref}
\usepackage{xcolor}
\usepackage{colortbl}
\definecolor{applegreen}{rgb}{0.55, 0.71, 0.0}
\definecolor{dodgerblue}{rgb}{0.12, 0.56, 1.0}
\definecolor{darkpastelgreen}{rgb}{0.01, 0.75, 0.24}
\definecolor{deepsaffron}{rgb}{1.0, 0.6, 0.2}
\definecolor{brandeisblue}{rgb}{0.0, 0.44, 1.0}
\definecolor{amaranth}{rgb}{0.9, 0.17, 0.31}

\def\eg{\emph{e.g}.} 
\def\ie{\emph{i.e}.} 
 
\def\etc{\emph{etc}.} 
\def\wrt{w.r.t.} 
\def\etal{\emph{et al}.}

\title{\LARGE \bf
Robust Self-Supervised LiDAR Odometry via Representative Structure Discovery and 3D Inherent Error Modeling 
}

\author{
Yan Xu,~~Junyi Lin,~Jianping Shi,~Guofeng Zhang,~Xiaogang Wang,~Hongsheng Li 
\thanks{Y. Xu, X. Wang, and H. Li are with the Department of Electronic Engineering, The Chinese University of Hong Kong, China. 
E-mail: {\tt yanxu@link.cuhk.edu.hk}, {\tt\{xgwang,hsli\}@ee.cuhk.edu.hk}}
\thanks{J. Lin and J. Shi are with SenseTime Research. 
E-mail: {\tt\{linjunyi,shijianping\}@sensetime.com}
} 
\thanks{G. Zhang is with the State Key Lab of CAD\&CG, Zhejiang University, China. E-mail: {\tt zhangguofeng@zju.edu.cn }
}%
}

\begin{document}

\maketitle
\thispagestyle{empty}
\pagestyle{empty}


\begin{abstract}
  The correct ego-motion estimation basically relies on the understanding of correspondences between adjacent LiDAR scans. However, given the complex scenarios and the low-resolution LiDAR, finding reliable structures for identifying correspondences can be challenging. In this paper, we delve into structure reliability for accurate self-supervised ego-motion estimation and aim to alleviate the influence of unreliable structures in training, inference and mapping phases. We improve the self-supervised LiDAR odometry substantially from three aspects: 1) A two-stage odometry estimation network is developed, where we obtain the ego-motion by estimating a set of sub-region transformations and averaging them with a motion voting mechanism, to encourage the network focusing on representative structures. 2) The inherent alignment errors, which cannot be eliminated via ego-motion optimization, are down-weighted in losses based on the 3D point covariance estimations. 3) The discovered representative structures and learned point covariances are incorporated in the mapping module to improve the robustness of map construction. Our two-frame odometry outperforms the previous state of the arts by $16\%$/$12\%$ in terms of translational/rotational errors on the KITTI dataset and performs consistently well on the Apollo-Southbay datasets. We can even rival the fully supervised counterparts with our mapping module and more unlabeled training data.

  \end{abstract}
  
  \section{Introduction}
\begin{figure}
    \centering
    \includegraphics[width=0.99\linewidth, trim=0.5cm 13.5cm 32cm 1cm, clip]{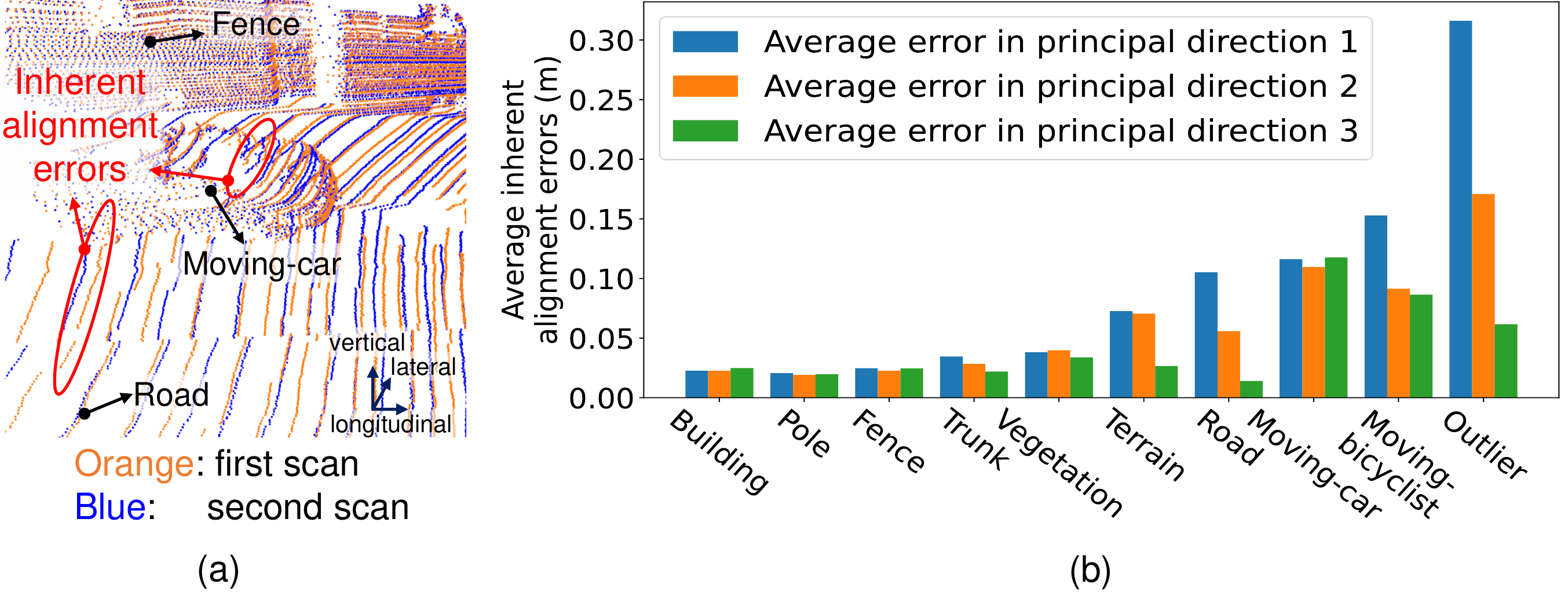}
    \vspace{-3ex}
    \caption{
      (a) 
      The complex 
      scenarios and low-resolution LiDAR increase the difficulty of correspondence identification for odometry. 
      Even after being transformed with ground-truth ego-motion, the 
      LiDAR scans can not be well aligned and the inherent alignment errors are inevitable. 
      (b) 
      We analyze the inherent errors for common semantic classes separately and re-project the errors via PCA.  The misalignment magnitude differs across classes and directions, reflecting a complex distribution of reliable regions 
      for finding correspondences. 
    }
    \label{fig:misalign}
    \vspace{-3ex}
\end{figure}

Ego-motion estimation from LiDAR sequences, namely LiDAR odometry, is essential for a wide range of related tasks including unmanned vehicle navigation, scene flow estimation, localization, range sensing \etc~\cite{gojcic2021weakly,xu2019depth,huang2021vs}.
Compared with visual cameras, LiDAR sensors can well capture the 3D structures under different lighting conditions, which is beneficial for odometry.
Nevertheless, the sparse and noisy nature of LiDAR data also poses great challenges in reliable correspondence identification for ego-motion estimation.
Even if we align the two adjacent LiDAR scans with ground-truth ego-motion, the misalignments still exist in some regions as shown in Fig. \ref{fig:misalign}a. The moving objects (\eg, moving cars), horizontal regions (\eg, road), and noisy outliers contribute to these inherent alignment errors. These inherent misalignments could influence the correspondence identification and could also impede the ego-motion optimization as these errors can never be eliminated by optimizing the ego-motion.
How to distinguish and focus on the representative structures
while adaptively downweight the unreliable regions is still an important problem in LiDAR odometry.

LiDAR odometry system generally contains two major components, \ie~the two-frame odometry module and the mapping module~\cite{zhang2014loam,shan2018lego,li2019net, liosam2020shan}. 
The two-frame odometry module estimates the ego-motion from two adjacent LiDAR scans, which is the foundation of the whole system.
The mapping module summarizes the previous scans to store a reliable map (scene model) for scan-to-map pose refinement. 
Classic registration-based methods \cite{arun1987least,segal2009generalized,serafin2015nicp,velas2016collar} have been widely used for two-frame odometry. 
These methods simply rely on nearest-neighbor search to identify correspondences between adjacent scans, which generally lack mechanisms to handle noisy depth measurements and moving objects. 
In recent years, following the learning-based visual odometry \cite{konda2015learning,wang2017deepvo,zhou2017unsupervised, li2019sequential,yang2020d3vo},   
learning-based LiDAR odometry methods \cite{velas2018cnn, li2019net, chounsupervised,xu2020selfvoxelo} have shown their competitiveness in both accuracy and real-time efficiency. In these works, the self-supervised methods \cite{chounsupervised,xu2020selfvoxelo} avoid the need for labeled data and are more practical for actual deployment.
Xu~{\etal} \cite{xu2020selfvoxelo} demonstrated the effectiveness of 3D CNN backbone in LiDAR odometry.
But similar to other learning-based methods, they fed LiDAR scans into an encoder network for ego-motion regression and do not have a design to explicitly enforce the network to adaptively focus on reliable regions.
Besides, the self-supervised methods generally leverage the geometric consistency between scans for ego-motion supervision. As shown in Fig. \ref{fig:misalign}a, the geometric consistency may not exactly hold in practice, and the inherent errors could impede the training process. Current methods estimate uncertainty scalars to model these inherent errors in category level, based on which to downweight the inherent errors in losses. However, as analyzed in Fig. \ref{fig:misalign}b, the inherent error magnitudes not only differ across different categories, but also vary in different directions. We argue that a more meticulous model for inherent errors is needed to differentiate the uncertainty differences in different directions during training.

In this work, we delve into the reliability of structures captured by LiDAR sensors for self-supervised LiDAR odometry.
For the two-frame odometry, unlike the previous works that view the LiDAR scan as a whole, 
we view the scan as a series of rectangular sub-region blocks, namely geometric units. We consider the reliability of different sub-regions and propose a two-stage ego-motion estimation pipeline to focus the network on representative structures. Concretely, we encode the overall 3D scene into feature volumes with 3D CNNs, where each feature vector represents one of the sub-region blocks.
Based on these high-dimensional features, we estimate the relative transformations of all geometric units in the first stage and average them to obtain the ego-motion based on a voting mechanism   
in the second stage to alleviate the influence from unreliable structures.

To encourage the voting to concentrate on reliable sub-regions in a self-supervised manner, we introduce a 3D uncertainty-aware geometric consistency loss to check the quality of current voted ego-motion.    
To handle the complex inherent errors exhibited in Fig. \ref{fig:misalign}b, we estimate point-wise covariances to describe the point uncertainties in different directions by assuming 3D points follow independent Gaussian distributions. Our geometric consistency loss is derived to maximize the likelihood of observed alignment errors, which can better differentiate the inherent errors induced by unreliable structures and the non-inherent errors induced by erroneous ego-motion estimations during training.

Based on the discovered representative structures and the point-wise covariance estimations, an uncertainty-aware mapping module is introduced for scan-to-map pose refinement.  
Unlike the conventional methods \cite{zhang2014loam,shan2018lego} that solely rely on a hand-crafted criterion to select the stable keypoints for scan-to-map pose refinement, we narrow down the key-point search space to the representative structures or robustness and efficiency.  
We also step further than Li {\etal} \cite{li2019net} by incorporating the point covariance estimations in the map construction to alleviate the fusion of unreliable points that may degrade the pose refinement. 

To conclude, our contributions are summarized as: 
    (1)~
    We propose a two-stage ego-motion estimation framework 
    to focus the network on the representative structures for two-frame LiDAR odometry. 
    (2)~
    We estimate point-wise covariances via a 3D CNN to model the point uncertainties,  
    based on which to better downweight the inherent errors 
    for the self-supervised losses.
    (3)~
    We introduce an uncertainty-aware mapping module that takes the discovered representative structures and the point-wise covariance estimations from CNNs
    for map construction and scan-to-map pose refinement. 
    (4)~
    Our self-supervised two-frame odometry estimation 
    outperforms the previous state of the arts 
    by $16\%$/$12\%$ in terms of translational/rotational error  on the KITTI dataset~\cite{geiger2012we} and performs consistently well on the Apollo-Southbay~\cite{lu2019l3}. With our mapping module and more training data, our system can even rival the supervised counterparts with real-time efficiency. Code is available at \url{https://github.com/SamuelYale/RSLO}.

  \section{Method}
  \begin{figure}[tb!]
    \centering
    \includegraphics[width=0.95\linewidth, trim=1cm 30cm 25.6cm 1.5cm, clip]{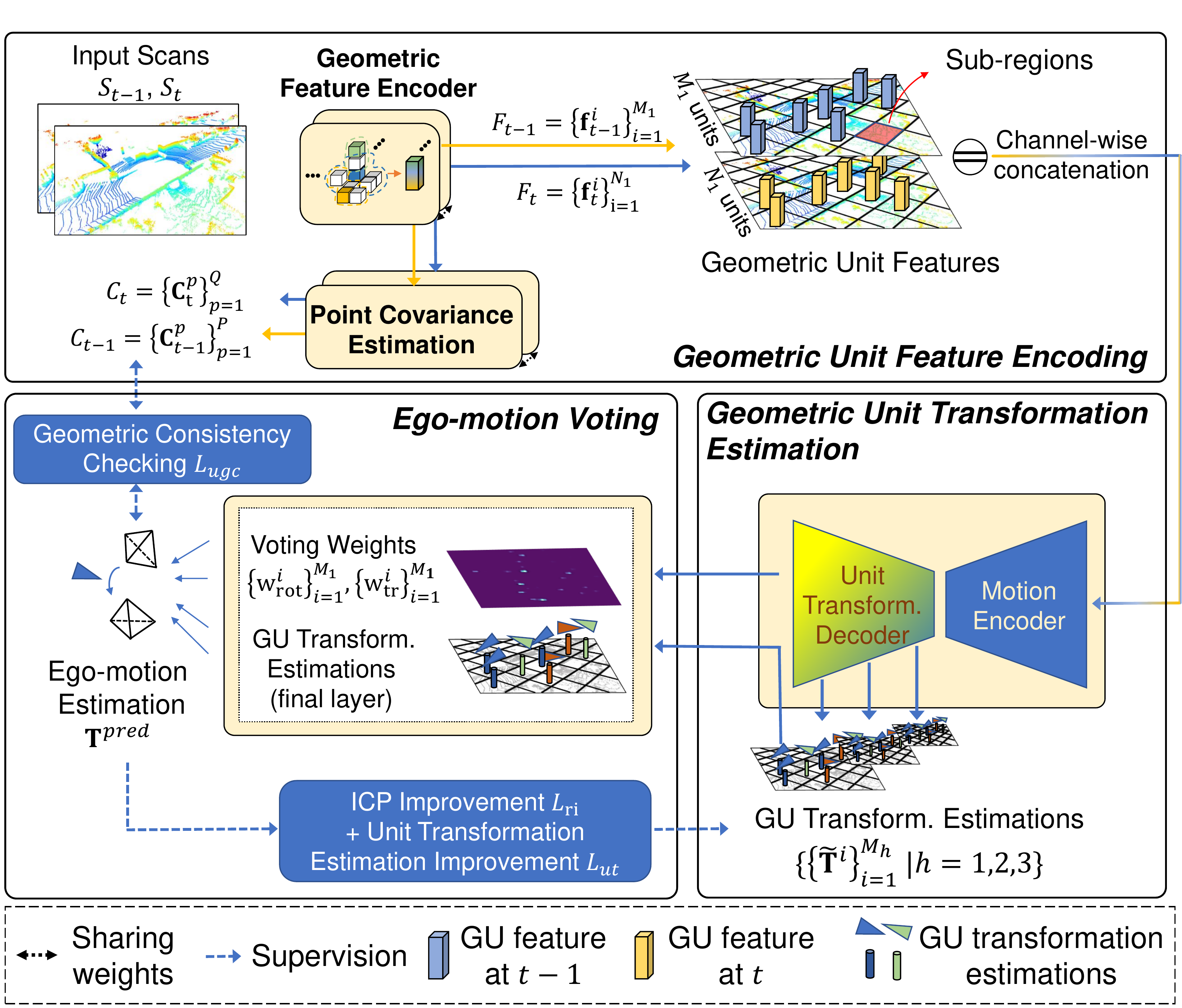}
    \vspace{-1ex}
    \caption{
      Our two-frame odometry network. 
      The geometric feature encoder encodes the input LiDAR scans $S_{t-1}$ and $S_{t}$ into 
      high-dimensional feature vectors $F_{t-1}$ and $F_{t}$. Each feature represents a sub-region block of the scan, and we name the sub-region as geometric unit (GU). 
      The channel-wise concatenation of these features are fed to the \textbf{geometric unit transformation estimation} module to estimate the geometric unit (GU) transformations $\{\widetilde{\mathbf{T}}^i\}_{i=1}^{M_h}$ between two scans hierarchically.
      During \textbf{ego-motion voting}, ego-motion $\mathbf{T}^{pred}$ is voted 
      based on the final-layer GU transformation estimations $\{\widetilde{\mathbf{T}}^i\}_{i=1}^{M_1}$. 
      The predicted voting weights
      would concentrate on the units with representative structures via our self-supervised training pipeline. 
    }
    \label{fig:framework}
    \vspace{-3ex}
\end{figure}

We propose a self-supervised LiDAR odometry system that counts the structure reliabilities in different phases. 
Specifically, 1) we propose a two-stage pipeline for the two-frame odometry inference (Sec. \ref{sec:lidar_odom_estimate}) to focus the network on representative structures; 2) we model the inherent alignment errors meticulously (Sec. \ref{sec:training pipeline}) to downweight them in loss functions; 3) we adopt a data-driven point reliability estimation strategy into the mapping module (Sec. \ref{sec:mapping}) for further robustness.
We will elaborate on each part in this section. 

\subsection{Two-frame LiDAR Odometry Estimation}\label{sec:lidar_odom_estimate}

Ego-motion measures the rigid transformation of an agent between consecutive timestamps.
The transformed scan points with the ground-truth ego-motion of one timestamp should perfectly align with the scan points at the other if there are no moving objects in the scene and the LiDAR sensor measures exactly the same spatial locations at the two timestamps.  
Most current works \cite{li2019net,chounsupervised, xu2020selfvoxelo} are based on the {rigid-scan-transformation} assumption and directly regress for the transformation of the whole scan. 
However, such rigid-scan-transformation assumption is generally not true in complex driving scenarios with non-ideal LiDAR as illustrated in Fig. \ref{fig:misalign}.
The non-rigid parts of the scan would inevitably affect the transformation estimation based on this assumption, as the transformations of non-rigid parts are not consistent with the ego-motion.  
To alleviate such dilemma, we propose to view the scan as a series of sub-regions and assume that only a portion of sub-regions corresponding to static objects with good measurement conditions undergo rigid transformations.  
Based on this relaxed assumption, we divide the ego-motion estimation pipeline into two stages, where the rigid transformations of sub-regions are estimated in the first stage, and these sub-region transformations are used to vote for the ego-motion with different weights in the second stage as shown in Fig. \ref{fig:framework}. This design can alleviate the influences of non-rigid parts and focus the network on sub-regions with representative structures. 
To achieve real-time efficiency, we design a  heterogeneous network that contains a 3D geometric feature encoder and a 2D geometric unit transformation estimation U-Net for sub-regions transformation estimation, which balances the speed and accuracy. For ego-motion voting, we adopt a self-attention-based voting module to identify the regions with representative structures and adjust the voting weights on different sub-regions. In the following parts of the sub-section, we will 
detail our two-stage odometry network.  

\subsubsection{Sub-region Representation}\label{sec:subregion_rep}
We adopt a 3D CNN constituted by submanifold convolution layers~\cite{graham2017submanifold} as the geometric feature encoder (shown in Fig. \ref{fig:framework})
to encode sub-region features.   
The geometric feature encoder efficiently processes whole scan point clouds 
$S_{t-1}$ and $S_{t}$ into high-dimensional features.   
The feature vectors in regular 3D convolution grids naturally relate to a rectangular sub-region block in the scan. We take the set of features output from the last convolution layer, \ie, $F_{t-1}$ and $F_{t}$, to represent different sub-regions, where 
each feature vector encodes both the sub-region geometry and the local contexts. 
We name these regular sub-regions as \textbf{geometric units} and the features as  \textbf{geometric unit features}.

\subsubsection{Geometric Unit Transformation Estimation}\label{sec:unit_transformation_estimate}
Explicitly identifying the geometric unit correspondence between two scans is non-trivial and inefficient. We hence adopt an efficient U-Net~\cite{ronneberger2015u} architecture as our geometric unit transformation estimation module to estimate all the geometric unit transformations in one shot. As shown in Fig. \ref{fig:framework}, firstly, we follow Xu {\etal }~\cite{xu2020selfvoxelo} to 
reshape the geometric unit features as bird-eye-view feature maps and then channel-wisely concatenate them, which keeps the spatial topology and the rigid motions. Next, we feed these features to the 
encoder-decoder network for geometric unit transformation estimation.

To make the geometric unit transformation estimation more locally-related, \ie, depending more on the local geometry and contexts,   
we estimate the geometric unit transformation in their self-centered coordinate frame $\widetilde{O}^i$.  
The conversion relation $\mathcal{T}$ between the rigid whole-scan transformation (also the ego-motion) $\mathbf{T}=\big(\begin{smallmatrix}
  \mathbf{R} & \mathbf{t}\\
  \mathbf{0} & 1
\end{smallmatrix}\big)\in SE(3)$  and a rigid sub-region transformation $\widetilde{\mathbf{T}}^i= \big(\begin{smallmatrix}
  \widetilde{\mathbf{R}}^i & \widetilde{\mathbf{t}}^i\\
  \mathbf{0} & 1
\end{smallmatrix}\big)\in SE(3)$ in $\widetilde{O}^i$ can be derived as 
\begin{equation}\label{eq:unit transformation}
  \mathcal{T}(\mathbf{T})
  =\begin{pmatrix}
      \mathbf{R} & \mathbf{t}+\mathbf{R}\mathbf{v}^i -\mathbf{v}^i\\
      \mathbf{0} & 1 
  \end{pmatrix}
  = \begin{pmatrix}
    \widetilde{\mathbf{R}}^i & \widetilde{\mathbf{t}}^i\\
    \mathbf{0} & 1 \\
    \end{pmatrix}
  =\widetilde{\mathbf{T}}^i,
\end{equation}
where $\mathbf{v}^i$ measures the offset between the geometric unit frame $\widetilde{O}^i$ and the LiDAR frame $O_L$. Fig. \ref{fig:self_centered} provides a further illustration on the relation between the geometric unit frame $\widetilde{O}^i$ and the LiDAR frame $O_L$.  
Here, we predict the rotation components $\widetilde{\mathbf{R}}^i$ in quaternion form for simplicity. To improve the training robustness, we hierarchically estimate the transformations of geometric units of different scales, \ie, $\{\{\widetilde{\mathbf{T}}^{i}\}_{i=1}^{M_h}| h=1,2,3\}$, from different depths of the unit transformation decoder as illustrated in Fig.~\ref{fig:framework}.  
But only the most accurate final-layer predictions  $\{\widetilde{\mathbf{T}}^{i}\}_{i=1}^{M_1}$ are used in the follow-up ego-motion voting. 
Here, $h$ indicates the scale level and $M_h$ denotes the geometric unit number in level $h$.

\subsubsection{Ego-motion Voting }\label{sec:eg_motion_vote}
If all geometric units are rigid and all the transformations are ideally estimated, accurate ego-motion can be estimated by simply averaging~\cite{gramkow2001averaging} over all the unit transformation estimations (after inversely converted with Eq.~\eqref{eq:unit transformation} to LiDAR coordinate system).
However, in practical scenarios, the non-rigid geometric units are ubiquitous, \eg, the units containing sparse/noisy measurements or moving objects.  
The transformation estimation of these regions is generally unstable or even ill-posed.   
The network should focus on the rigid geometric units of which transformations can be well estimated while ignoring the non-rigid ones. 
\begin{figure}[b!]
    \vspace{-3ex}
    \centering
    \begin{subfigure}[t]{0.53\linewidth}
        \includegraphics[width=0.95\linewidth, trim=0.3cm 16cm 45cm 0.7cm, clip]{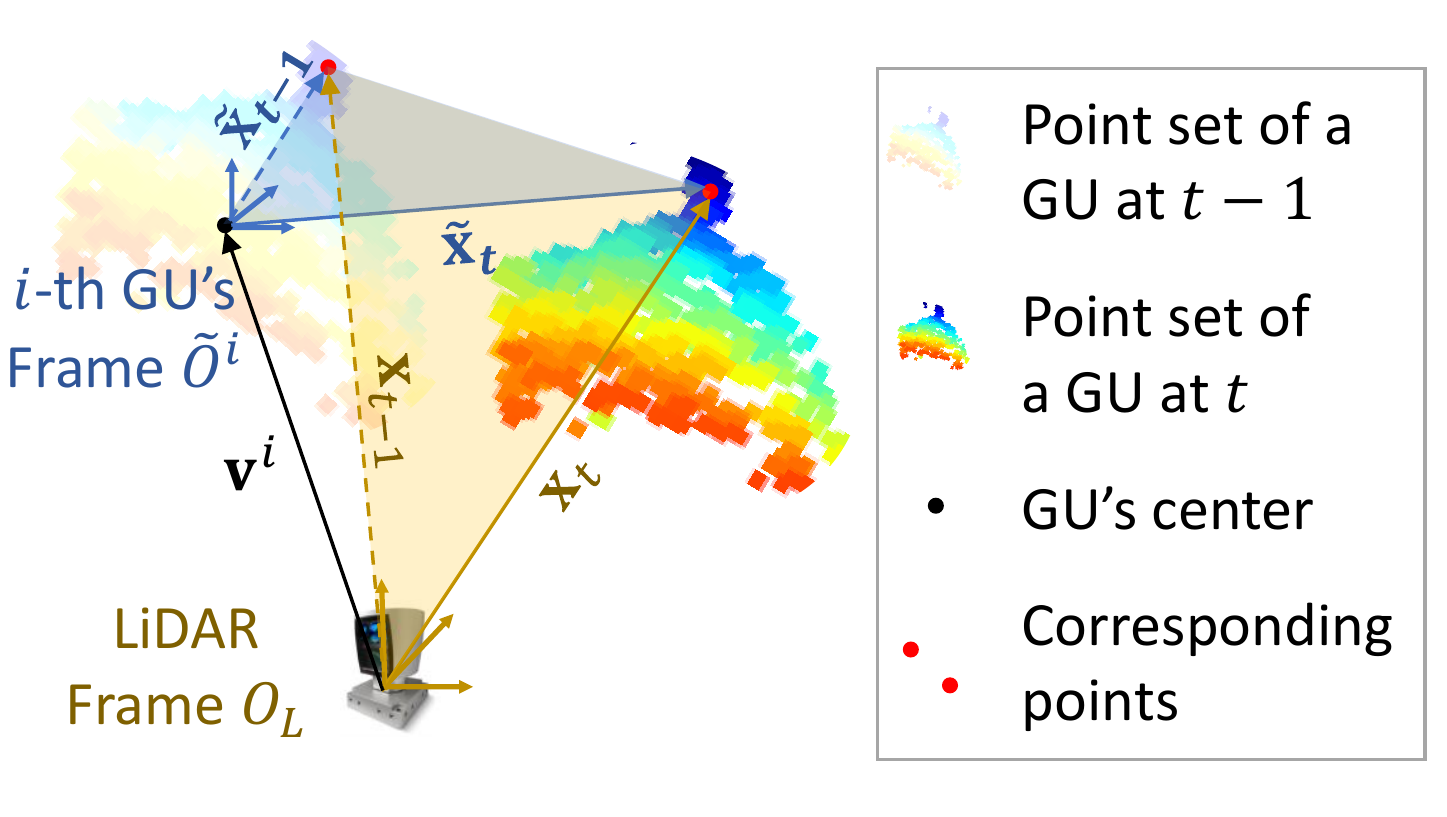}
    \vspace{-3ex}
    \caption{}\label{fig:self_centered}
    \end{subfigure}
\hfill
\begin{subfigure}[t]{0.45\linewidth}
    \centering
    \includegraphics[width=1\linewidth, trim=4.5cm 17.5cm 45.5cm 0.7cm, clip]{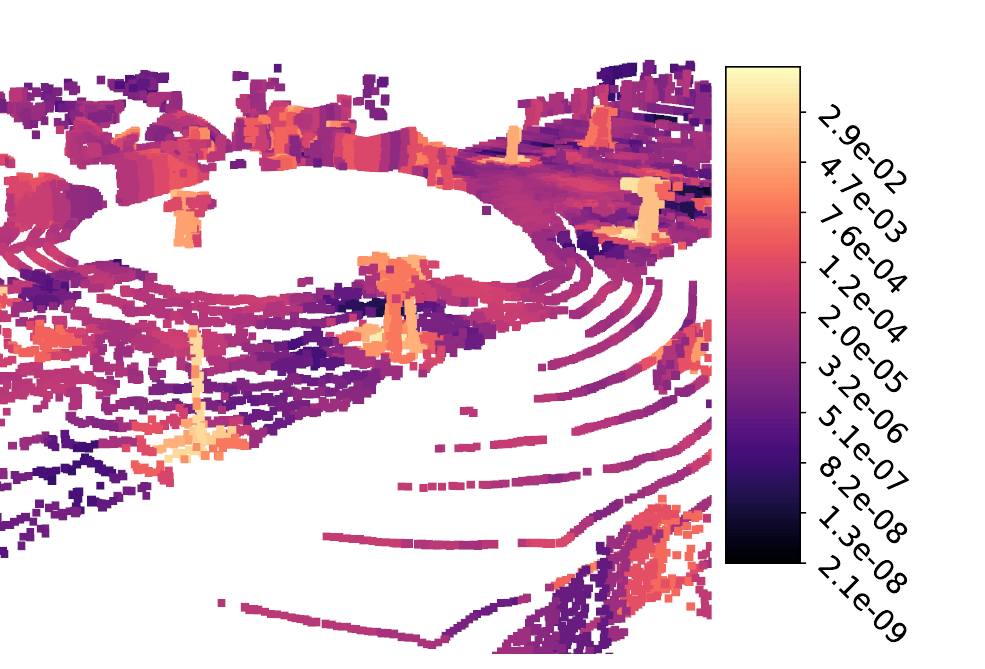}
    \caption{}\label{fig:rep_structure}
\end{subfigure}
    \vspace{-1ex}
    \caption{(a) 
    Demonstration of the relation between a geometric unit (GU) frame and the LiDAR frame.  
    (b)~Visualization of the voting weights on different scan sub-regions.  
    Brighter color denotes larger weights. 
    }
    
    \vspace{-1ex}
\end{figure}

    
We therefore propose a self-attention based ego-motion voting module (shown in Fig. \ref{fig:framework}) to find
reliable geometric units containing representative structures and focus the network on them for ego-motion estimation. 
Based on the last feature map of the decoder in previous geometric unit transformation estimation module, the voting module predicts 
the selection scores $\{l_{rot}^{i}\}_{i=1}^{M_1}$ and $\{l_{tr}^{i}\}_{i=1}^{M_1}$ for different units regarding to the \textbf{rot}ation and \textbf{tr}anslation estimations separately. 
The selection scores are normalized with a standard softmax function over all geometric units to be the voting weights of different geometric units: 
$\{w_{rot}^{i}\}_{i=1}^{M_1}$ and $\{w_{tr}^i\}_{i=1}^{M_1}$. 
With the voting weights, 
the ego-motion $\mathbf{T}^{pred}$ is estimated by weightedly averaging the geometric unit transformations $\{\widetilde{\mathbf{T}}^{i}\}_{i=1}^{M_1}$. 
Before the voting, unit transformation $\widetilde{\mathbf{T}}^i$ is inversely converted via Eq. \eqref{eq:unit transformation}, \ie, $\mathcal{T}^{-1}(\widetilde{\mathbf{T}}^i)$,  to get the rotation component $\mathbf{R}^i$ and transformation component $\mathbf{t}^i$ in the LiDAR coordinate frame, and the voting process can then be expressed as  
\begin{equation}\label{eq:top_vote}
  \mathbf{T}^{pred} = \begin{pmatrix}
      \sum_{i=1}^{M_1} w_{rot}^i\mathbf{R}^i & \sum_{i=1}^{M_1} w_{tr}^i\mathbf{t}^i \\
      \mathbf{0} & 1
  \end{pmatrix}.  
\end{equation} 
Note that the weighted average of $\mathbf{R}^i$ \cite{gramkow2001averaging} is based on the quaternion representations with a slight abuse of notation.
Trained with our self-supervised losses detailed in the next section, the voting module is able to 
well distinguish the confident units and assign them voting weights orders of magnitude larger than others. Fig. \ref{fig:rep_structure} visualizes the voting weight predictions.

\subsection{Self-supervised Training with 3D Inherent Alignment Error Modeling}\label{sec:training pipeline}

\subsubsection{3D uncertainty-aware Geometric Consistency Loss}\label{sec:ugc}
To avoid the need for ground-truth ego-motion, current self-supervised methods \cite{zhou2017unsupervised,chounsupervised, xu2020selfvoxelo} leverage the geometric consistency between adjacent scans for ego-motion training.  
Basically, scan $S_t$ at timestamp $t$ is first transformed with the ego-motion estimation to align it with the previous scan $S_{t-1}$. 
Then, the discrepancy between the transformed scans serves as a proxy loss to supervise the ego-motion estimation. 
However, the geometric consistency may not exactly hold between scans, as the measurements of LiDAR are sparse and the surrounding environments are complex.  
As exhibited in Fig. \ref{fig:misalign}, we could easily identify the inherent alignment errors by aligning two adjacent scans with accurate ego-motion. 
These inherent errors might impede the training if we do not differentiate them from the non-inherent errors caused by erroneous ego-motion estimations.  
From Fig. \ref{fig:misalign}b, we could observe that the inherent alignment error distributions 1) not only differ across object categories (\eg, moving objects and noisy outliers have larger errors)
2) but also vary in different directions:  \eg, for road and terrain, the error magnitudes in the third principal direction are much smaller than those in the first two principal directions\footnote{The third principal direction roughly coincides with the vertical direction, while the other two principal directions spread in the ground plane.}. 
Existing methods~\cite{zhou2017unsupervised,li2019net,chounsupervised, xu2020selfvoxelo} estimate point-wise uncertainty scalars to model the inherent errors in \textit{category level}, based on which certain types of regions are ignored
during optimization. 
We argue that the geometric information may not be fully exploited for odometry if we do not consider the differences of measurement reliability in directions. For example, the category-level uncertainty modeling could lower the importance of road regions during training due to the observable inherent errors along the longitudinal direction, as shown in Fig. \ref{fig:misalign}a. However, the road regions are important in rectifying the erroneous vertical motion estimations \cite{shan2018lego}, as the road regions even have smaller average inherent error magnitude in the vertical direction. 
We hence may need to estimate the measurement uncertainties in different directions to alleviate such dilemma. 

To this end, we associate a $3\times 3$ covariance matrix $\mathbf{C}_t^p$ with each point to describe its uncertainty in 3D space,  by assuming that each scan point $\mathbf{x}_{t}^p\in S_t$ follows an independent \textit{Gaussian distribution}, \ie, $\mathbf{x}_{t}^p \sim\mathcal{N}(\hat{\mathbf{x}}_t^p, \mathbf{C}_t^p)$, where $\hat{\mathbf{x}}_t^p$ denotes the ideal measurement. 
We use a 3D CNN (point covariance estimation module in Fig. \ref{fig:framework}) to estimate the covariances\footnote{We estimate the related non-negative eigenvalues and eigenvector matrix (corresponding to a normalized 4-dimensional vector) in practice to keep the positive semi-definite property of covariance matrix. }, $\mathbf{C}_{t-1}^p$ and $\mathbf{C}_{t}^p$, 
of every point in adjacent scans $S_{t-1}$ and $S_t$. 
We derive a loss function to guide the covariance learning and supervise the ego-motion estimation $\mathbf{T}^{pred}$ in an uncertainty-aware manner. 
Concretely, with the above assumption, the point alignment error (the difference between a transformed point $\mathbf{T}^{pred}\mathbf{x}_{t}^p$, and its nearest-neighbor $\mathbf{x}_{t-1}^p$ in previous scan $S_{t-1}$) also follows a Gaussian distribution: $\mathbf{e}_p=\mathbf{x}_{t-1}^p-\mathbf{T}^{pred}\mathbf{x}_{t}^p \sim \mathcal{N}(0, \mathbf{C}_{t-1}^p+\mathbf{T}^{pred} \mathbf{C}_{t}^p \mathbf{T}^{{pred}^T} )$. 
By letting $\mathbf{\Sigma}_p =\mathbf{C}_{t-1}^p+\mathbf{T}^{pred} \mathbf{C}_{t}^p \mathbf{T}^{{pred}^T}$, the negative log-likelihood of observed point alignment errors can thus be derived as   
\begin{equation}\label{eq:alignment_loss}
  L_{ugc}=\sum\limits_{p=1}^{|S_t|} \underbrace{\frac{1}{2}\mathbf{e}_{p}^{T} \mathbf{\Sigma}_{p}^{-1} \mathbf{e}_p}_\text{uncertainty-aware L2}   + \underbrace{\frac{1}{2}\log(\det(\mathbf{\Sigma}_p))}_\text{covariance regularization}, 
\end{equation}
which is taken as the uncertainty-aware geometric consistency loss to supervise the ego-motion estimation and the point covariances. The covariance regularization terms help to avoid trivial covariance estimations, \ie, producing all-zero weights $\mathbf{\Sigma}_p^{-1}$, and also push the network to reduce the weights on inherent errors in certain directions/regions while emphasizing the non-inherent errors in uncertainty-aware L2 terms for a smaller overall loss value. 
We empirically find that the network can model the covariance parameters very well given its powerful regression ability and large-scale training data. 


Classic methods~\cite{segal2009generalized} handcraft the point covariances as diagonal matrices with small eigenvalues along surface normals, which consider less on the practical point distributions. While the scalar uncertainty modeling adopted by previous methods can be seen as a special case of our Eq.~\eqref{eq:alignment_loss} with $\mathbf{\Sigma}_p$ set to $diag(\sigma,\sigma,\sigma)$,  where $\sigma$ is a scalar uncertainty estimation. 
By comparison, our model can handle the inherent errors more meticulously, as our weighting factor $\mathbf{\Sigma}_p$ can more flexibly adjust the weights on errors in different directions.  
We found this loss can more stably push the ego-motion estimation $\mathbf{T}^{pred}$ towards the final target and the covariance estimations $\mathbf{C}_t^p$ trained with this loss can also improve the mapping in Sec.~\ref{sec:mapping}. 
Moreover, this loss also pushes the ego-motion voting module to increase the voting weights of geometric units that contain representative structures helpful for transformation reasoning.

\subsubsection{Other Losses}

To improve the geometric unit transformation estimations in a self-supervised manner, we follow Xu \etal~\cite{xu2020selfvoxelo} to adopt a residual improvement loss $L_{ri}$ and generate a more accurate ego-motion $\mathbf{T}^*$ with ICP.  The ICP is initialized with the current ego-motion estimation $\mathbf{T}^{pred}$  and conducted incompletely (2 iterations) for efficiency. 
After obtaining the improved ego-motion target $\mathbf{T}^*$,  
we first approximate the geometric unit transformation targets $\{\{\widetilde{\mathbf{T}}^{*^i}\}_{i=1}^{M_h} | h=1,2,3\}$ for different scale levels $h$ with $\mathbf{T}^*$ via Eq.~\eqref{eq:unit transformation}. Then we supervise the geometric unit transformation estimations $\{\{\widetilde{\mathbf{T}}^{i}\}_{i=1}^{M_h}| h=1,2,3\}$ 
with these approximated targets:
\begin{equation}\label{eq:unit transformation loss}
  \small
   L_{ut}^h =   
   u_{\alpha}(\sum\limits_{i=1}^{M_h} {w}_{tr}^{i}||\widetilde{\mathbf{t}}^i-\widetilde{\mathbf{t}}^{*^i} ||^2_2) 
        +u_{\beta}(\sum\limits_{i=1}^{M_h}{w}_{rot}^{i}  || \widetilde{\mathbf{R}}^i-\widetilde{\mathbf{R}}^{*^i}||^2_2). 
\end{equation}
Here, the translation components $\widetilde{\mathbf{t}}^{i}$ and rotation components $\widetilde{\mathbf{R}}^i$ (in quaternion) of $\widetilde{\mathbf{T}}^{i}$ are supervised separately with the respective components $\widetilde{\mathbf{t}}^{*^i}$ and $\widetilde{\mathbf{R}}^{*^i}$ of target $\widetilde{\mathbf{T}}^{*^i}$. 
The robust losses $u_\alpha(\circ)=e^{-\alpha}\circ+\alpha$ and $u_\beta(\circ)=e^{-\beta}\circ+\beta$ are used \cite{kendall2017geometric}, where $\alpha$  and $\beta$ are learnable parameters. 
The focusing weights ${w}_{rot}^{i}$ and ${w}_{tr}^{i}$ are adjusted for different geometric units to 
concentrate the network on regions with representative structures.  
For the scale level $h=1$, these focusing weights are obtained via softmax normalization (with temperature parameter $\gamma>1$) on the selection scores generated during ego-motion voting (Sec. \ref{sec:eg_motion_vote}):  
${w}_{rot}^{i}=softmax(l_{rot}^{i}, \gamma)$ and ${w}_{tr}^{i}=softmax(l_{tr}^{i}, \gamma)$, 
where $l_{rot}^{i}$ and $l_{tr}^{i}$ are the selection scores.  
For other scale levels ($h=2,3$), we spatially downsample these weights by average pooling to focus on the corresponding parts in respective scale levels.   
The temperature parameter $\gamma>1$ is used to tune the flatness of the softmax function. A proper $\gamma$ value can provide valid supervision to more potentially reliable geometric units and encourage the network to identify more representative structures. 
We balance the losses of different scales 
in the final unit transformation loss $L_{ut}=\sum_{h=1}^3 \mathrm{w}_h L_{ut}^h$, where the balance weights $\mathrm{w}_h$ are empirically set as $0.5,0.25,0.1$ for $h=1,2,3$, respectively.

\subsection{Uncertainty-aware Mapping}\label{sec:mapping}
Although with the uncertainty modeling, the noisy and sparse nature of the LiDAR point cloud limits the accuracy of the two-frame-based odometry.
Existing methods~\cite{zhang2014loam,shan2018lego,li2019net} proposed to map the scene by accumulating the previous scans and further refine the ego-motion estimation by scan-to-map alignment. 
As mentioned before, the non-rigid parts and noisy measurements cause inconsistency among frames, which should be excluded from the scan-to-map alignment and map construction. 
Most of the existing methods adopt handcrafted criteria to select reliable parts for map construction and scan-to-map alignment, which is not always faithful, considering 
the pervasive unreliable points in practice. 
We handle this dilemma by incorporating the learning-based point covariance estimations in Sec. \ref{sec:ugc} 
and the priors provided by discovered representative structures into the mapping module.  


In our implementation, we use regular voxels to represent the global map, where each voxel has a coordinate $\overline{\mathbf{x}}_{t}^p\in\mathbb{R}^3$ and covariance $\overline{\mathbf{C} }^{p}_{t}\in\mathbb{R}^{3\times 3}$ to describe the voxel state at current timestamp $t$. The map is constructed by the incremental update that adds new LiDAR scan points to the existing map.  
Before each map update, we refine the two-frame ego-motion estimation $\mathbf{T}^{pred}$ by scan-to-map alignment optimization. The ego-motion is refined as $\mathbf{T}^{map}$ by 
first matching the scan points with the previous map $\mathcal{M}_{t-1}$ and then solving a nonlinear least squares problem, 
following Zhang \etal~\cite{zhang2014loam}.
Unlike Zhang~\etal~ that solely rely on handcrafted criteria to select keypoints in the entire new scan to match with the map, we narrow down the keypoint searching space to the representative structure set 
identified by criterion $w_{rot}^iw_{tr}^i>\tau_1$ to improve the robustness and efficiency, where  $w_{rot}^i$ and $w_{tr}^i$ are the 
learnt ego-motion voting weights of geometric units in Sec.~\ref{sec:eg_motion_vote}. 

With the refined pose $\mathbf{T}^{map}$, we transform the 
point cloud $\{\mathbf{x}_t^p\}_{p=1}^Q$ and the estimated point covariances $\{\mathbf{C}_t^p\}_{p=1}^Q$ 
of the new scan to the map coordinates and update the map with them. 
For uncertainty-aware map updating, we consider the reliabilities of both the incoming points and the old map voxels to improve the robustness. 
Specifically, if the transformed new point 
(\ie, with states  $\mathbf{x}_t^{p'} =\mathbf{T}^{map}\mathbf{x}_t^{p}$ and $\mathbf{C}_t^{p'} = \mathbf{T}^{map} \mathbf{C}_{t}^p \mathbf{T}^{{map}^T}$)
fall in empty spaces, we will create a new voxel there and directly set its state with the state of the new point: $\overline{\mathbf{x}}_{t}^p = \mathbf{x}_t^{p'}$ and $\overline{\mathbf{C}}_{t}^p=\mathbf{C}_t^{p'}$. 
Otherwise, we will leverage the Bayes filter \cite{barfoot2019state} for voxel update: 
\begin{equation}\label{eq:map_update}
  \small
  \begin{cases}
  &\overline{\mathbf{C} }^{p}_t = (\overline{\mathbf{C} }^{{p^{-1}} }_{t-1} +\mathbf{C}^{p'^{-1}}_{t})^{-1} \\
  &\overline{\mathbf{x}}^{p}_t = \overline{\mathbf{C} }^{p}_t (\overline{\mathbf{C} }^{{p^{-1} }}_{t-1} \overline{\mathbf{x}}_{t-1}^p  + \mathbf{C}^{p'^{-1}}_{t} {\mathbf{x}}^{p'}_{t} ). 
  \end{cases}
\end{equation}
We empirically find that our uncertainty-aware map construction can alleviate the fusion of unreliable points that have adverse influence in the scan-to-map pose refinement and lead to more robust performance.

  \section{Experiment}
  \subsection{Dataset, Evaluation Metrics and Experimental Setup}
\begin{table}[t] 
    \caption{Comparison with state-of-the-art methods on KITTI odometry dataset. 
    We compare with other competitive classic methods, supervised methods and unsupervised methods to fully evaluate our performance. 
    }
    \label{tab:eval_kitti}
      \centering
      \setlength\tabcolsep{2.5 pt}
      \resizebox{1\linewidth}{!}{
      \begin{tabular}{r|c|cccccccccccc}
        \hline
        &\multirow{2}{*}{Seq.} 
        &\multicolumn{2}{c|}{7}&
        \multicolumn{2}{c|}{8}
        &\multicolumn{2}{c|}{9}
        &\multicolumn{2}{c|}{10} 
        &\multicolumn{2}{c}{Avg.} \\
        \cline{3-12}
        &&\multicolumn{1}{c}{$t_{rel}$}&$r_{rel}$ 
        &$t_{rel}$&$r_{rel}$
        &$t_{rel}$&$r_{rel}$
        &$t_{rel}$&$r_{rel}$
        &$t_{rel}$&$r_{rel}$ \\
        \hline
        \multirow{7}{*}{\rotatebox{90}{Classic} } &ICP-po2po 
        &5.17 &3.35 & 10.04&4.93&6.93&2.89&8.91&4.47 &7.76&3.98\\\arrayrulecolor{gray}\cline{2-12}
        &ICP-po2pl 
        &1.55 &1.42 & 4.42&2.14&3.95&1.71&6.13&2.60&4.01&1.97 \\\cline{2-12}
       
        &{CLS~\cite{velas2016collar}} 
        &1.04 &0.73 & 2.14&1.05&1.95&0.92&3.46&1.28&2.15&1.00 \\\cline{2-12}
        &{NDT-P2D~\cite{stoyanov2012fast}} 
        &7.51&3.07&13.6&4.62&33.7&7.06&20.5&3.06&18.8&4.45\\\cline{2-12}
        &{GICP~\cite{segal2009generalized}} 
        &0.64 &0.45 & 1.58&0.75&1.97&0.77&1.31&0.62&1.38&0.65\\\cline{2-12}
        &{LOAM~\cite{zhang2014loam}} &0.69 &0.50 & 1.18&0.44&1.20&0.48&1.51&0.57&1.15&0.50 \\\cline{2-12}
        &{LeGO-LOAM~\cite{shan2018lego}} &1.20 &0.59 & 1.59&0.61&1.51&0.63&1.99&0.68&1.57&0.63 \\\cline{2-12}
        \arrayrulecolor{black} \hline
       
        \multirow{5}{*}{\rotatebox{90}{Sup.} }
        &{Velas~\etal~\cite{velas2018cnn}}  
        & 2.89& - &4.94& - &3.27&-&3.22&- \\\arrayrulecolor{gray}\cline{2-12}
        &{\begin{tabular}{@{}c@{}} LO-Net~\\(w/o map)~\cite{li2019net} \end{tabular}}  
        &1.70 &0.89 & 2.12&{0.77}&1.37&0.58&1.80&0.93&1.75&0.79 \\\cline{2-12}
        &{\begin{tabular}{@{}c@{}} LO-Net~\\(w/ map)~\cite{li2019net} \end{tabular}}  
        &0.56 &0.45 & {1.08}&0.43&0.77&0.38&{0.92}&0.41&{0.83}&0.42 \\\cline{2-12}
        \arrayrulecolor{black} \hline
        \multirow{14}{*}{\rotatebox{90}{Unsup.} }
        &Zhou~\etal*~\cite{zhou2017unsupervised}  
        &21.3 &6.65 & 21.9&2.91&18.8&3.21& 14.3&3.30&19.1&4.02 \\ \arrayrulecolor{gray}\cline{2-12} 
        &UnDeepVO*~\cite{li2018undeepvo}  
        &3.15 &2.48 & 4.08&1.79&7.01&3.61& 10.6&4.65&6.22&3.13 \\ \cline{2-12}
        &{Cho~\etal~\cite{chounsupervised}}  
        &-&- & -& - &4.87&1.95& 5.02&1.83&4.95&1.89 \\ \cline{2-12}
        &{ \begin{tabular}{@{}c@{}} Xu~\etal~\\(w/o map)~\cite{xu2020selfvoxelo}\end{tabular} }
        &3.09&1.81&3.16 & 1.14 &3.01&1.14& 3.48&1.11&3.19&1.30 \\ \cline{2-12}
        &{\begin{tabular}{@{}c@{}} Xu~\etal~(+data, \\w/o map)~\cite{xu2020selfvoxelo}\end{tabular}} 
        &2.51&1.15&2.65 & 1.00 &2.86&1.17& 3.22&1.26&2.81&1.15  \\\cline{2-12} 
        &{\begin{tabular}{@{}c@{}} Xu~\etal~(+data, \\w/ map)~\cite{xu2020selfvoxelo}\end{tabular}} 
        &$\mathbf{0.34}$&$\mathbf{0.21}$&{1.18} & $\mathbf{0.35}$ &{0.83}&{0.34}& {1.22}&{0.40}&{0.89}&{0.32} \\ \arrayrulecolor{black} \cline{2-12} \noalign{\vskip\doublerulesep\vskip-\arrayrulewidth}\cline{2-12}
        &Ours (w/o map) 
        &3.24&1.72&2.48 &1.10  &2.75&1.01& 3.08&1.23&2.89&1.23 \\ \arrayrulecolor{gray}\cline{2-12}
        &\begin{tabular}{@{}c@{}} Ours~(+data, \\w/o map) \end{tabular}
        &2.37&1.15& 2.14& 0.92 &2.61&1.05& 2.33&0.94&2.36&1.02 \\ \cline{2-12}
        &\begin{tabular}{@{}c@{}} Ours~(+data, \\w/ map) \end{tabular}
        &0.56&{0.26}&\textbf{1.17} & {0.38} &\textbf{0.65}&\textbf{0.25}& \textbf{0.72}&\textbf{0.31}&\textbf{0.78}&\textbf{0.31} \\ 
        \arrayrulecolor{black}\hline 
        
      \end{tabular}
      }
  
      {
        \raggedright 
        \scriptsize
        ~~$t_{rel}$ and $r_{rel}$ are the average translational RMSE (\%) and rotational RMSE ($^\circ$/100m) on lengths of 100m-800m~\cite{geiger2012we}, and * denotes visual odometry methods. `more data' indicates using sequences  (00-06, 11-22) for training, otherwise, the training sequences are (00-06). Some results are from \cite{li2019net}.
      \par}
      \vspace{-3ex}
  \end{table}

\textbf{Dataset}.
We evaluate our method on two large-scale datasets, KITTI Odometry dataset \cite{geiger2012we} and Apollo SouthBay Dataset~\cite{lu2019l3}, 
captured by cars driven at varying speeds in country roads, urban areas, highways \etc

\textbf{Evaluation Metrics.}
We adopt official evaluation metrics in KITTI benchmark~\cite{geiger2012we}, which measure the translational and rotational drifts on lengths of 100m-800m.

\textbf{Experimental Details}
Our two-frame odometry network is implemented with PyTorch and trained on NVIDIA Tesla V100 GPUs. 
The point clouds are voxelized as voxels of sizes (0.1m,0.1m,0.2m) before input to the network.
We adopt an initial learning rate of 0.001 and slowly decay it with a cosine annealing strategy.
We take 3 adjacent LiDAR frames to form temporal pairs  
$[S_{t-2}, S_{t-1}]$, $[S_{t-1}, S_t]$, $[S_{t-2}, S_t]$ following~\cite{li2019net}, and then randomly sample 16 such tuples to create a batch for each training iteration. The network is trained with 200k iterations for all the experiments.
The weights of losses $L_{ugc}$, $L_{ri}$ and $L_{ut}$ are set to 1, and the temperature parameter $\gamma$ is set to $20$ in $L_{ut}$.
Note that, in the first training epoch, we warm up the network by supervising with identity transformations to avoid divergence.  
For the representative structure selection, the threshold $\tau_1$ is set to the $60$-th percentile of $\{w_{rot}^i\cdot w_{tr}^i\}$. For the mapping module, we adopt a leaf voxel size of 0.8m following LOAM~\cite{zhang2014loam}.

  \begin{table}[t] 
    \caption{Evaluation results on Apollo-SouthBay test set.}
    \vspace{-1ex}
    \label{tab:eval_apollo}
    \begin{center}
      \resizebox{0.5\linewidth}{!}{
      \setlength\tabcolsep{1.5 pt}
      \begin{tabular}{c|cccccccccccccccccccccccc}
        \hline
         &Avg.~$t_{rel}$& Avg.~$r_{rel}$ \\
        \hline\hline
        {ICP-po2po} &22.8 &2.35  \\
        {ICP-po2pl}  & 7.75&1.20 	\\
        {GICP~\cite{segal2009generalized}}  & 4.55	& 0.76\\
        {NDT-P2D~\cite{stoyanov2012fast}} &57.2 & 9.40 \\
        {LOAM~\cite{zhang2014loam}} &5.93 &0.26 \\
        {Xu \etal~\cite{xu2020selfvoxelo} (w/o map) }  &6.42&1.65\\
        {Xu \etal~\cite{xu2020selfvoxelo} (w/ map) }& 2.25&0.25\\
        \hline 
        {Ours (w/o map)}  &5.99&1.58 \\
        {Ours (w/ map)} &$\mathbf{2.17}$&$\mathbf{0.24}$\\

        \hline 
      \end{tabular}
      }
  
      \vspace{-6ex}
    \end{center}
  \end{table}

\begin{figure*}[ht]
    \centering
    \begin{subfigure}[t]{0.5\linewidth}
    \centering
    \includegraphics[width=0.9\linewidth, trim=1cm 13cm 33.5cm 1cm, clip]{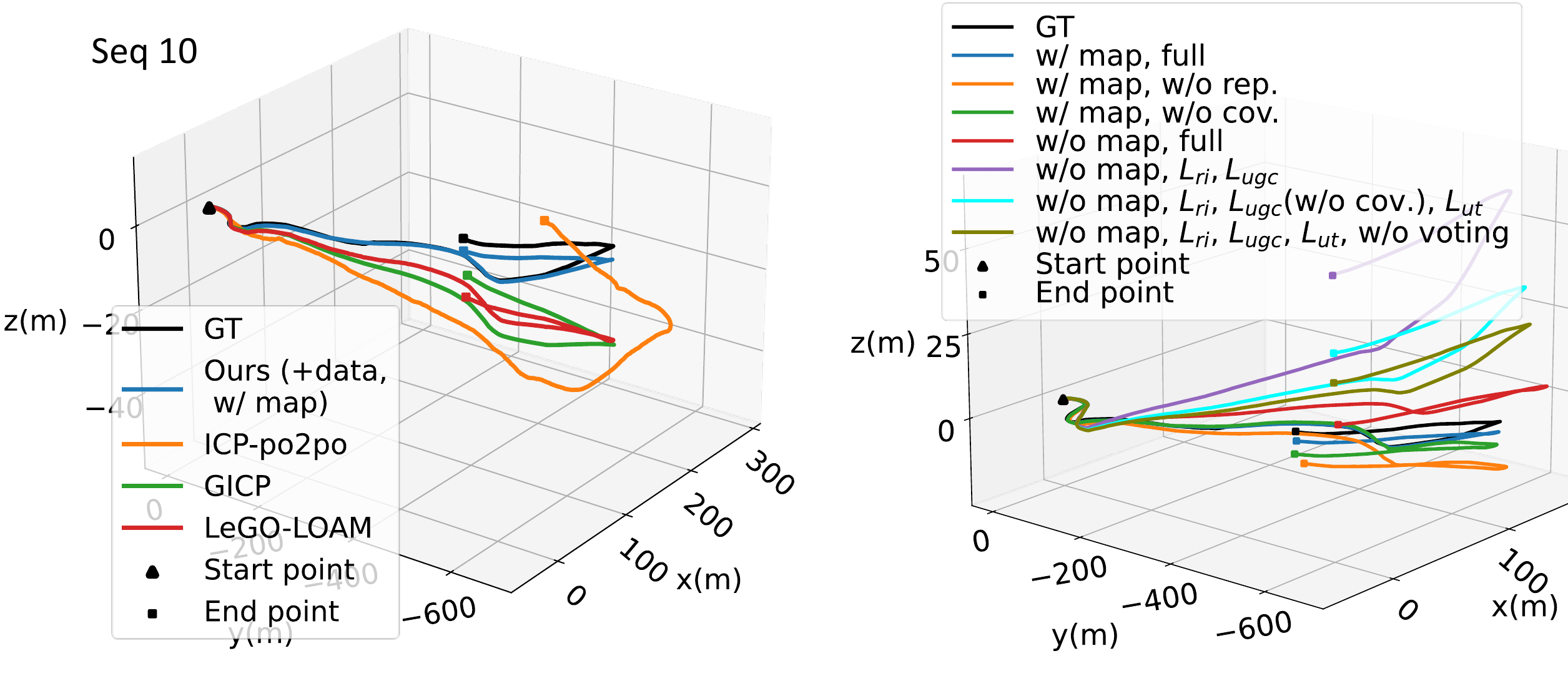}
    \vspace{-1.6ex}
    \caption{}\label{fig:traj_plot}
    \end{subfigure}
    \hspace{-1.4ex}
    \begin{subfigure}[t]{0.28\linewidth}
      \centering
    \includegraphics[width=\linewidth, trim=2cm 1.3cm 42cm 13cm, clip]{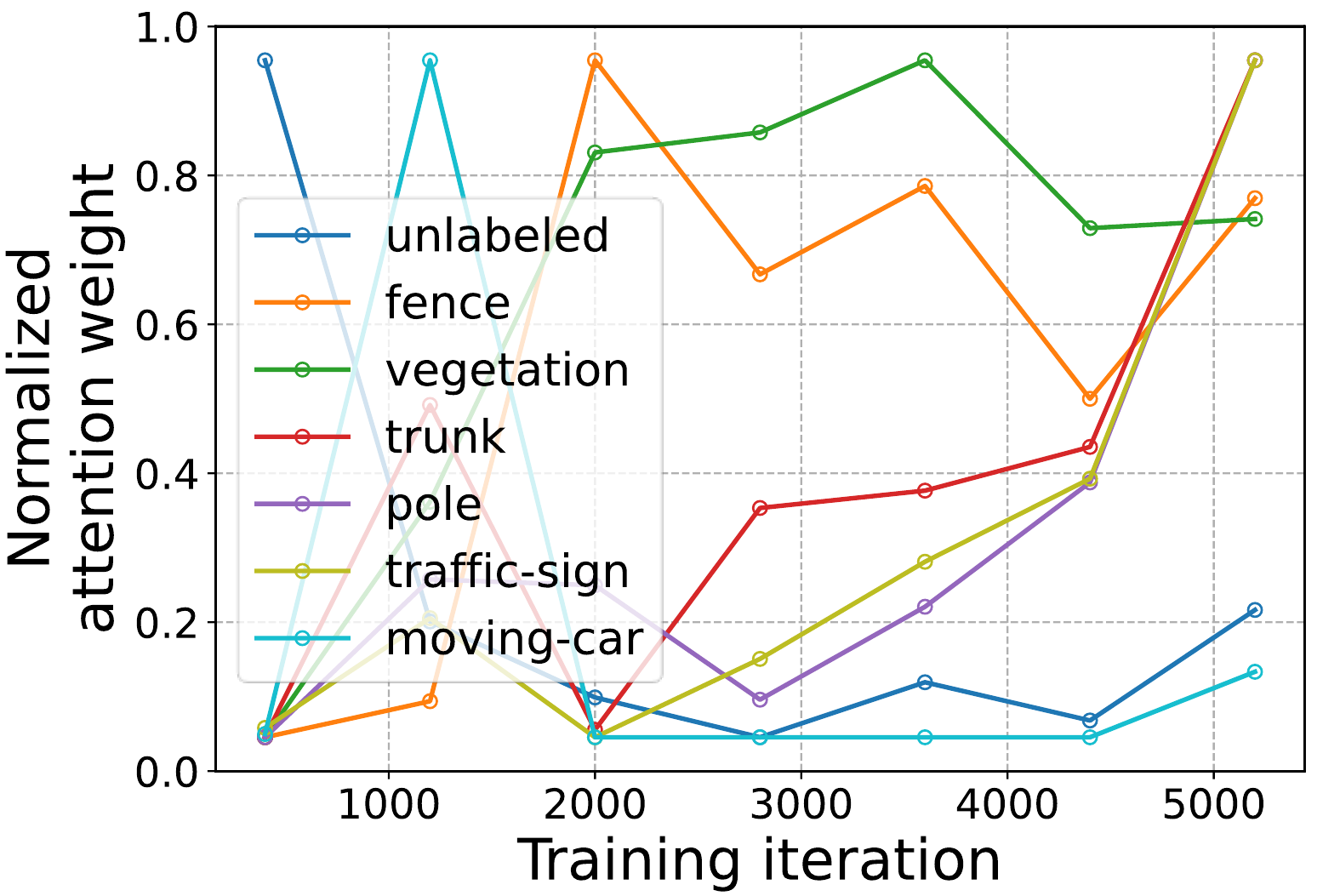}
    \vspace{-4ex}
    \caption{}\label{fig:weight_converge}
    \end{subfigure}
    \begin{subfigure}[t]{0.15\linewidth}
     \centering
    \includegraphics[width=\linewidth, trim=0cm 2cm 42.5cm 0.5cm, clip]{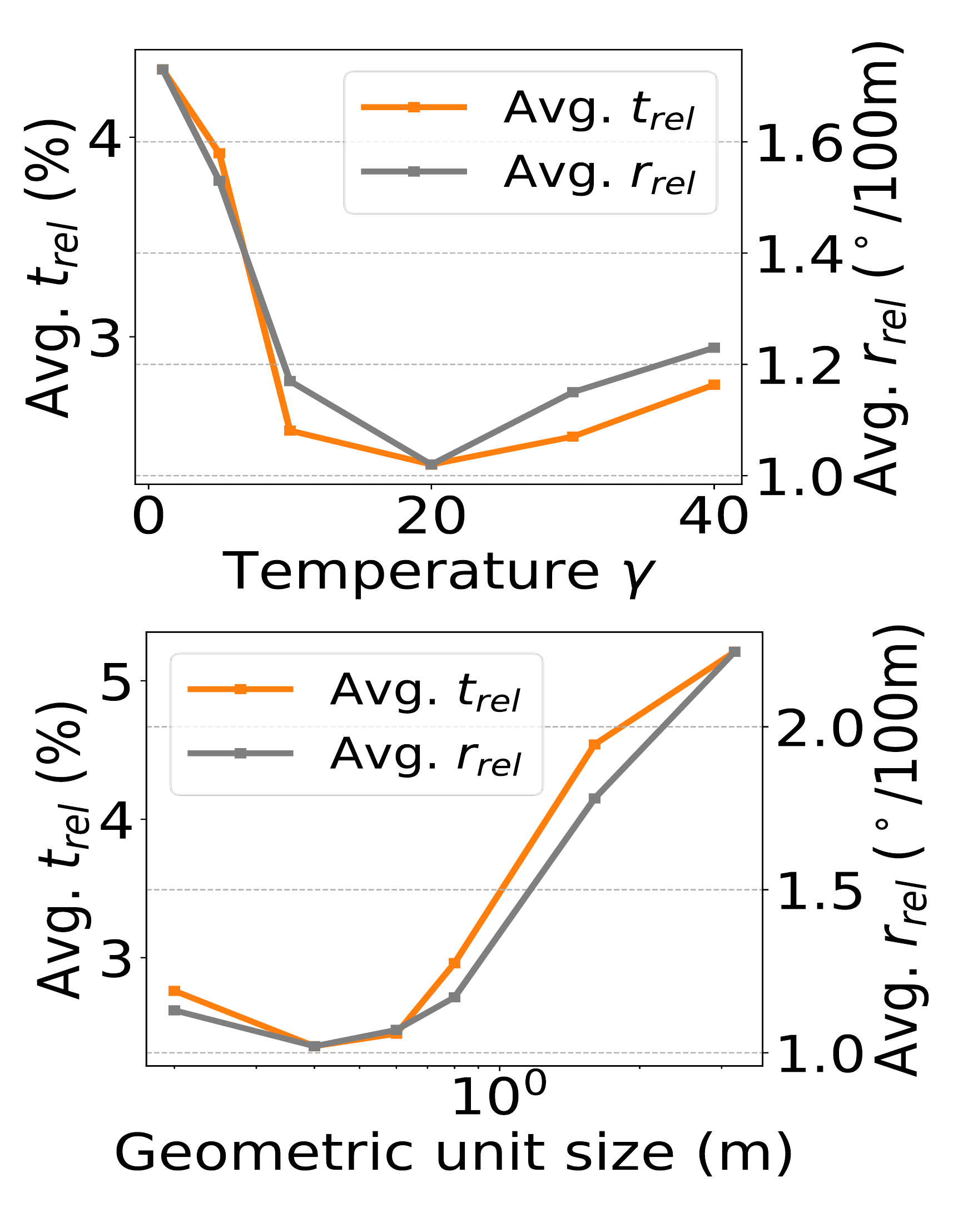}
    \vspace{-4.2ex}
    \caption{}\label{fig:param}
    \end{subfigure}
    \vspace{-1.5ex}
    \caption{
      (a) The 3D visualization of estimated trajectories by different methods (right) and different ablation variants (left)  
      on KITTI sequence 10. The variants in the right figure correspond to those in Table~\ref{tab:ablation}. 
    (b) The convergence of the 
    ego-motion voting weights
    during training.
    (c) The performance variation with different temperature $\gamma$ values in unit transformation loss (Eq.~\eqref{eq:unit transformation loss}) or different geometric unit sizes.
    }
    \vspace{-3ex}
\end{figure*}
\begin{figure}[tb]
    \centering
    \includegraphics[width=0.99\linewidth, trim=0.5cm 10.5cm 22.5cm 0cm, clip]{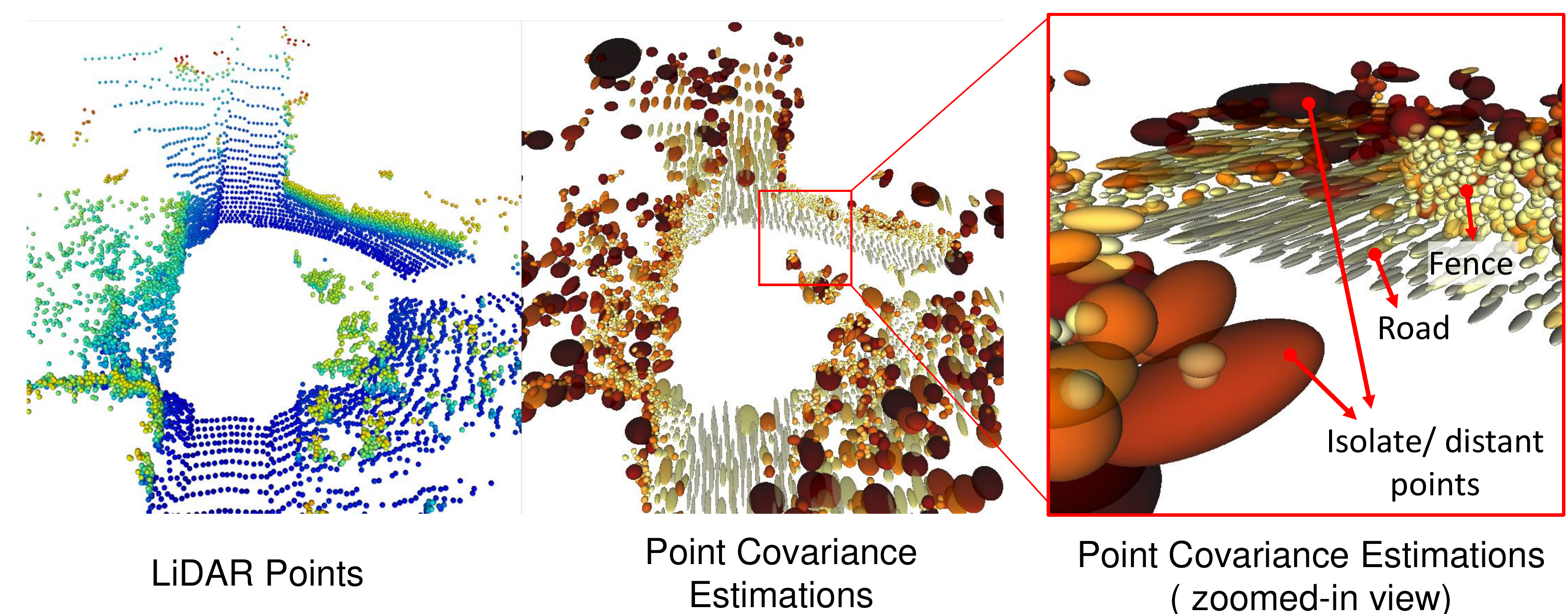}
    \caption{
    Visualization of point covariance estimations. 
    We decompose each point covariance to three orthogonal directions via eigendecomposition and visualize the variance magnitudes as lengths of the orthogonal ellipsoid axes. 
    The ellipsoid is colorized according to the smallest variance magnitude among 
    three orthogonal directions.  
    Lighter color denotes smaller magnitude and higher reliability.  
    }
    \label{fig:elips_cov_vis}
    \vspace{-4ex}
\end{figure}
\subsection{Comparison with State-of-the-Art Methods}
\textbf{Evaluation on KITTI Odometry Dataset.}
We compare our method with state-of-the-art methods including both the classic methods and the learning-based methods. 
For the classic methods, we choose some competitive point registration methods \ie, ICP~\cite{arun1987least}, CLS~\cite{velas2016collar}, NDT-P2D~\cite{stoyanov2012fast}, GICP~\cite{segal2009generalized}, and the LiDAR SLAM system LOAM~\cite{zhang2014loam} and LeGO-LOAM~\cite{shan2018lego} for comparison. 
For the learning-based methods, we select the cutting-edge supervised methods~\cite{velas2018cnn,li2019net} and the unsupervised methods~\cite{chounsupervised,xu2020selfvoxelo} as baselines.
Different learning-based methods may have different dataset splits for training and testing. For example, Velas \etal~\cite{velas2018cnn} take the seq. 00-07 for training and seq.~08-10 for testing, Li \etal~\cite{li2019net,xu2020selfvoxelo} take seq.~00-06 for training and seq.~07-10 for testing while the others take 00-08~/~09-10 as training/testing splits.
Considering most of their codes are not available, for a fair comparison, 
we follow \cite{xu2020selfvoxelo} to adopt two different splitting strategies for training/testing, \ie, 00-06/07-10 and (00-06,11-21)/07-10, to evaluate the effectiveness of our method. 
As shown in Table~\ref{tab:eval_kitti}, our self-supervised two-frame odometry estimation (denoted as `Ours (w/o map)') outperforms the cutting-edge self-supervised counterparts (Cho~\etal~\cite{chounsupervised} and Xu \etal (w/o map) \cite{xu2020selfvoxelo} by large margins with the same or even less amount of data for training. 
By adding more training data (using seq. 00-06 and 11-21 for training), our two-frame odometry performance (denoted as `Ours (+data, w/o map)') can be further improved by $18.3\%$ (2.36 vs. 2.89) and $17.1\%$ (1.02 vs. 1.23) in terms of translational and rotational errors, which also outperforms the state of the art (`Xu \etal~(+data, w/o map)') with the same settings.
With more training data, our two-frame based odometry also outperforms most classic point registration methods, \ie, point-to-point ICP \cite{arun1987least}, point-to-plane ICP~\cite{chen1992object} and NDT-P2D \cite{stoyanov2012fast}.
Although the CLS~\cite{velas2016collar} and GICP~\cite{segal2009generalized} achieve more robust performance, they need time-consuming iterative refinement and are not suitable for real-time systems. 

The last row in Table~\ref{tab:eval_kitti} (`Ours (+data, w/ map)') shows our overall performance with the proposed uncertainty-aware mapping module, which achieves the best average performance among compared methods.
With discovered representative structures and  estimated point covariances, our scan-to-map pose refinement robustly improves the ego-motion estimated by our two-frame odometry network. 
Note that, with more training data, our self-supervised method achieves comparable results to the state-of-the-art supervised counterpart (`LO-Net (w/ map)'). Moreover, Fig.~\ref{fig:traj_plot} visualizes the trajectories from different methods. Our method consistently performs well and achieves low drift errors even after traversing long distances.

\textbf{Evaluation on Apollo-SouthBay Dataset.} 
We further test the robustness of our method on  Apollo-SouthBay.
Although the Apollo dataset is more complicated and the driving distances are much longer than KITTI, our method still robustly achieves accurate odometry estimation. Table~\ref{tab:eval_apollo} lists our experimental results. Our two-frame odometry estimation (denoted as `Ours (w/o map)') 
still performs well and even achieves comparable translation estimation performance to that of the state-of-the-art LiDAR SLAM system LOAM~\cite{zhang2014loam}. Our overall system with the uncertainty-aware mapping module (denoted as `Ours (w/ map)') achieves the best odometry performance among all the compared classic methods and the recent learning-based method~\cite{xu2020selfvoxelo}.

\subsection{Effectiveness of Proposed Key Components}
\begin{table}[t] 
  \caption{Effects of individual components in our proposed framework. The models are trained with sequences (00-06, 11-21) for this table. 
  }
 \vspace{-2ex}
  \label{tab:ablation}
  \setlength\tabcolsep{2.5 pt}
  \begin{center}
    \resizebox{0.68\linewidth}{!}{
    \begin{tabular}{c|c|cc}
      \hline
      &Test &Avg. $t_{rel}$ &Avg. $r_{rel}$ \\
      \hline
      \hline
      \multirow{7}{*}{\rotatebox{90}{w/o map} }&$L_{ri}, L_{ugc}$ 
      &4.24&1.64\\
      &$L_{ri}, L_{ugc}(\mathrm{w/o~cov}), L_{ut}$ 
      &3.03&1.20\\
      &$L_{ri}, L_{ugc}(\mathrm{w/~scalar~conf.}), L_{ut}$
      &2.93&1.15\\ 
      &$L_{ri}, L_{ugc}, L_{ut}$, {w/o~voting} 
      &2.72&1.21\\ 
      &$L_{ri}, L_{ugc}, L_{ut}$, {w/ 7D-voting} 
      &2.63&1.09\\ 
      &$L_{ri}, L_{ugc}, L_{ut}, att.$ &3.66&1.48\\
      &$L_{ri}, L_{ugc}, L_{ut}$ (full) 
      &$\mathbf{2.36}$&$\mathbf{1.02}$\\ 
      \hline
      \multirow{3}{*}{\rotatebox{90}{w/ map} }& w/ map, w/o~rep. 
      &0.88&0.32 \\ 
      &w/ map, w/o~cov. 
      &0.85&0.35 \\ 
      &w/map, full
      &$\mathbf{0.78}$&$\mathbf{0.31}$\\ 
      \hline 
    \end{tabular}
    }
    \vspace{-6ex}
  \end{center}
\end{table}

      

The ablation study is conducted on the KITTI dataset to verify the effectiveness of our proposed components. The results are listed in Table~\ref{tab:ablation}. 

\textbf{Losses in self-supervised training pipeline}.
We try to train the network without the unit transformation loss $L_{ut}$ (denoted as `$L_{ri},L_{ugc}$'), and find the performance degrades significantly, because most of the geometric unit transformations cannot be estimated well without this supervision and thus the ego-motion voting module do not have enough good candidates to produce accurate estimations.   
We also verify the effectiveness of our uncertainty-aware geometric consistency loss $L_{ugc}$ (Eq.~\eqref{eq:alignment_loss}), by replacing the covariance components $\mathbf{\Sigma}_p$ with identity matrices to make $L_{ugc}$ a standard L2 loss 
(denoted as `$L_{ri}, L_{ugc} (\mathrm{w/o~cov}), L_{ut}$'), or using scalar confidences to replace it (denoted as `$L_{ri}, L_{ugc} (\mathrm{w/~scalar~conf.}), L_{ut}$').  
These ablation variants do not perform as well as our full model (denoted as `$L_{ri}, L_{ugc}, L_{ut}$ (full)') due to their less considerate modeling for inherent alignment errors during training.
To give a better understanding of the effect of our point covariance estimation, we visualize the estimated covariances in Fig.~\ref{fig:elips_cov_vis}. 
We can observe that the nearby planar regions usually have small variance magnitudes. For the ground points, ``flat-and-slim'' covariance estimations indicate the ground points have higher uncertainties along the longitudinal direction. This is reasonable because the low-resolution LiDAR usually produces observable measurement errors in this direction, as visualized in Fig.~\ref{fig:misalign}a.  Besides, for noisy points and remote areas, the estimated variances are usually significant in all directions.

\textbf{With or without ego-motion voting.}
We show the necessity of ego-motion voting by assigning equal voting weights to all the geometric units, of which results correspond to the `$L_{ri}, L_{ugc}, L_{ut}$, w/o~voting' in Table \ref{tab:ablation}. 
Influenced by the unstable estimations of  unreliable geometric units, the performance is downgraded as expected.
We further statistically analyze the convergence of ego-motion voting weight estimation on some typical categories annotated by SemanticKITTI~\cite{behley2019iccv}
and observe reasonable phenomena. From Fig.~\ref{fig:weight_converge}, we can see that the voting weights are quickly concentrated on some representative structures like fences, poles, and trunks during training, while the weights on moving objects and unlabeled noisy regions gradually decrease as the training proceeds.  
We further test the voting by using 7-dimensional covariance (denoted as `$L_{ri}, L_{ugc}, L_{ut}$, {w/ 7D-voting}') with the similar spirit of \cite{richter2019towards}, but find no further improvement. We deem that our one-dimensional weight for rotation/translation could be more compact to hold the high-level information (\ie, representative or not), and could be more predictable by our network. 
Besides, we also tried some attention modules \cite{vaswani2017attention} in network layers, but found little effects in our system (denoted by `$L_{ri}, L_{ugc}, L_{ut}, att.$'). 

\textbf{Effectiveness of representative-structure prior and uncertainty-aware map construction in mapping module.} 
We further check the effect of representative-structure selection (by voting weights) and the effect of our uncertainty-aware map construction in our mapping module.  
The variant `w/ map, w/o~rep.' in Table \ref{tab:ablation} does not limit the keypoint searching space within discovered representative structures,   and directly search planar and edge features from the whole scan for scan-to-map pose refinement.  
The performance degrades since the unreliable points in the new scan would more likely to be used for the scan-to-map alignment optimization without the priors.  
Moreover, we also try to abandon the proposed   uncertainty-aware map construction (Eq.~\eqref{eq:map_update}) and take an average filter to fuse all new points equally for map construction. The less accurate results of this ablation version, \ie, `w/ map, w/o~cov.', demonstrate the necessity of our uncertainty-aware map update.

\subsection{Analysis of Convergence and Stability }

We analyze the stability {\wrt } the hyperparameters.  As shown in Fig.~\ref{fig:param}, too small or too large temperatures $\gamma$ in $L_{ut}$ (Eq.~\eqref{eq:unit transformation loss}) degrades the performance.  Because a too small $\gamma$ value limits the valid supervision scopes to only a few units,  which impedes the geometric unit transformation learning. 
While with a too large $\gamma$, the supervision on non-rigid units would also influence the training adversely, because the conversion relation of Eq.~\eqref{eq:unit transformation} does not hold for these units, and thus the approximate targets are wrong.
Besides, the geometric unit size also affects the performance as shown in Fig.~\ref{fig:param} (below). Too small or too large unit sizes could degrade the performance, as too small unit sizes are not suitable for geometric feature extraction while too large unit sizes would decrease the effect of ego-motion voting.

\vspace{-1ex}
\subsection{Runtime Analysis}
We test the running time on a machine with an Intel(R) Core(TM) i7-9700K CPU and a NVIDIA Tesla V100 GPU.
In our system, the learning-based two-frame odometry is running on a GPU and the scan-to-map pose refinement is conducted on another thread in the backend on a CPU. As shown in Table~\ref{tab:runtime}, our method achieves real-time efficiency, which is suitable for practical deployment. 

    

\begin{table}[t]
  \centering
   \caption{The runtime of our submodules.}\label{tab:runtime}
   \vspace{-1ex}
  \resizebox{0.8\linewidth}{!}{
    \begin{tabular}{c*{1}{|c}}
   \hline
     Module & Time (ms) \\
     \hline
     \hline
     Geometric unit feature encoding & 74.7\\
     Geometric unit transfromation estimation+Ego-motion voting & 23.2 \\
     Mapping (in the backend) & 73.4 \\
     \hline
     Overall &  97.9\\
     \hline
   \end{tabular}
   }
   \vspace{-5ex}
\end{table}

\section{Conclusions}
We present a self-supervised LiDAR odometry system, including a two-frame odometry network and an uncertainty-aware mapping module. The two-frame odometry network first estimates geometric unit transformations from high-dimensional 3D features, and then votes for the ego-motion based on the discovered representative structures. The uncertainty-aware mapping module constructs a reliable map with the point-wise covariance estimations from the 3D CNN and refines the two-frame-based ego-motion estimation with previously discovered representative structures. Experiments on public datasets demonstrate the effectiveness of our method and the proposed system is able to run in real-time. 

\vspace{1ex}
\noindent\textbf{Acknowledgement}
{
\linespread{-1}
\footnotesize
Yan thanks all the coauthors' efforts in improving this paper. Also, he sincerely expresses his appreciation for the valuable feedback from Li Yang and Bangbang Yang during the manuscript revision. 
}






\bibliographystyle{IEEEtran}
\bibliography{IEEEabrv,mybib}

\begin{thebibliography}{10}
\providecommand{\url}[1]{#1}
\csname url@rmstyle\endcsname
\providecommand{\newblock}{\relax}
\providecommand{\bibinfo}[2]{#2}
\providecommand\BIBentrySTDinterwordspacing{\spaceskip=0pt\relax}
\providecommand\BIBentryALTinterwordstretchfactor{4}
\providecommand\BIBentryALTinterwordspacing{\spaceskip=\fontdimen2\font plus
\BIBentryALTinterwordstretchfactor\fontdimen3\font minus
  \fontdimen4\font\relax}
\providecommand\BIBforeignlanguage[2]{{%
\expandafter\ifx\csname l@#1\endcsname\relax
\typeout{** WARNING: IEEEtran.bst: No hyphenation pattern has been}%
\typeout{** loaded for the language `#1'. Using the pattern for}%
\typeout{** the default language instead.}%
\else
\language=\csname l@#1\endcsname
\fi
#2}}

\bibitem{gojcic2021weakly}
Z.~Gojcic, O.~Litany, A.~Wieser, L.~J. Guibas, and T.~Birdal, ``Weakly
  supervised learning of rigid 3d scene flow,'' in \emph{Proceedings of the
  IEEE/CVF Conference on Computer Vision and Pattern Recognition}, 2021, pp.
  5692--5703.

\bibitem{xu2019depth}
Y.~Xu, X.~Zhu, J.~Shi, G.~Zhang, H.~Bao, and H.~Li, ``Depth completion from
  sparse lidar data with depth-normal constraints,'' in \emph{Proceedings of
  the IEEE/CVF International Conference on Computer Vision}, 2019, pp.
  2811--2820.

\bibitem{huang2021vs}
Z.~Huang, H.~Zhou, Y.~Li, B.~Yang, Y.~Xu, X.~Zhou, H.~Bao, G.~Zhang, and H.~Li,
  ``Vs-net: Voting with segmentation for visual localization,'' in
  \emph{Proceedings of the IEEE/CVF Conference on Computer Vision and Pattern
  Recognition}, 2021, pp. 6101--6111.

\bibitem{zhang2014loam}
J.~Zhang and S.~Singh, ``Loam: Lidar odometry and mapping in real-time.'' in
  \emph{Robotics: Science and Systems}, vol.~2, no.~9, 2014.

\bibitem{shan2018lego}
T.~Shan and B.~Englot, ``Lego-loam: Lightweight and ground-optimized lidar
  odometry and mapping on variable terrain,'' in \emph{2018 IEEE/RSJ
  International Conference on Intelligent Robots and Systems (IROS)}, 2018, pp.
  4758--4765.

\bibitem{li2019net}
Q.~Li, S.~Chen, C.~Wang, X.~Li, C.~Wen, M.~Cheng, and J.~Li, ``Lo-net: Deep
  real-time lidar odometry,'' in \emph{Proceedings of the IEEE Conference on
  Computer Vision and Pattern Recognition}, 2019, pp. 8473--8482.

\bibitem{liosam2020shan}
T.~Shan, B.~Englot, D.~Meyers, W.~Wang, C.~Ratti, and R.~Daniela, ``Lio-sam:
  Tightly-coupled lidar inertial odometry via smoothing and mapping,'' in
  \emph{IEEE/RSJ International Conference on Intelligent Robots and Systems
  (IROS)}.\hskip 1em plus 0.5em minus 0.4em\relax IEEE, 2020, pp. 5135--5142.

\bibitem{arun1987least}
K.~S. Arun, T.~S. Huang, and S.~D. Blostein, ``Least-squares fitting of two
  3-{D} point sets,'' \emph{{IEEE} Trans. Pattern Anal. Mach. Intell.}, vol.~9,
  no.~5, pp. 698--700, 1987.

\bibitem{segal2009generalized}
A.~Segal, D.~Haehnel, and S.~Thrun, ``Generalized-icp.'' in \emph{Robotics:
  science and systems}, vol.~2, no.~4.\hskip 1em plus 0.5em minus 0.4em\relax
  Seattle, WA, 2009, p. 435.

\bibitem{serafin2015nicp}
J.~Serafin and G.~Grisetti, ``Nicp: Dense normal based point cloud
  registration,'' in \emph{2015 IEEE/RSJ International Conference on
  Intelligent Robots and Systems (IROS)}, 2015, pp. 742--749.

\bibitem{velas2016collar}
M.~Velas, M.~Spanel, and A.~Herout, ``Collar line segments for fast odometry
  estimation from velodyne point clouds,'' in \emph{2016 IEEE International
  Conference on Robotics and Automation (ICRA)}, 2016, pp. 4486--4495.

\bibitem{konda2015learning}
K.~R. Konda and R.~Memisevic, ``Learning visual odometry with a convolutional
  network.'' in \emph{VISAPP (1)}, 2015, pp. 486--490.

\bibitem{wang2017deepvo}
S.~Wang, R.~Clark, H.~Wen, and N.~Trigoni, ``Deepvo: Towards end-to-end visual
  odometry with deep recurrent convolutional neural networks,'' in \emph{2017
  IEEE International Conference on Robotics and Automation (ICRA)}, 2017, pp.
  2043--2050.

\bibitem{zhou2017unsupervised}
T.~Zhou, M.~Brown, N.~Snavely, and D.~G. Lowe, ``Unsupervised learning of depth
  and ego-motion from video,'' in \emph{Proceedings of the IEEE Conference on
  Computer Vision and Pattern Recognition}, 2017, pp. 1851--1858.

\bibitem{li2019sequential}
S.~Li, F.~Xue, X.~Wang, Z.~Yan, and H.~Zha, ``Sequential adversarial learning
  for self-supervised deep visual odometry,'' in \emph{Proceedings of the IEEE
  Int. Conf. on Computer Vision}, 2019, pp. 2851--2860.

\bibitem{yang2020d3vo}
N.~Yang, L.~v. Stumberg, R.~Wang, and D.~Cremers, ``{D3VO}: Deep depth, deep
  pose and deep uncertainty for monocular visual odometry,'' in
  \emph{Proceedings of the IEEE/CVF Conference on Computer Vision and Pattern
  Recognition}, 2020, pp. 1281--1292.

\bibitem{velas2018cnn}
M.~Velas, M.~Spanel, M.~Hradis, and A.~Herout, ``Cnn for imu assisted odometry
  estimation using velodyne lidar,'' in \emph{2018 IEEE International
  Conference on Autonomous Robot Systems and Competitions (ICARSC)}, 2018, pp.
  71--77.

\bibitem{chounsupervised}
Y.~Cho, G.~Kim, and A.~Kim, ``Unsupervised geometry-aware deep lidar
  odometry,'' in \emph{2020 International Conference on Robotics and Automation
  (ICRA)}.\hskip 1em plus 0.5em minus 0.4em\relax IEEE, 2020.

\bibitem{xu2020selfvoxelo}
Y.~Xu, Z.~Huang, K.-Y. Lin, X.~Zhu, J.~Shi, H.~Bao, G.~Zhang, and H.~Li,
  ``{SelfVoxeLO}: Self-supervised lidar odometry with voxel-based deep neural
  networks,'' in \emph{Conference on Robot Learning}, 2020.

\bibitem{geiger2012we}
A.~Geiger, P.~Lenz, and R.~Urtasun, ``Are we ready for autonomous driving? the
  kitti vision benchmark suite,'' in \emph{2012 IEEE Conference on Computer
  Vision and Pattern Recognition}, 2012, pp. 3354--3361.

\bibitem{lu2019l3}
W.~Lu, Y.~Zhou, G.~Wan, S.~Hou, and S.~Song, ``L3-net: Towards learning based
  lidar localization for autonomous driving,'' in \emph{Proceedings of the IEEE
  Conference on Computer Vision and Pattern Recognition}, 2019, pp. 6389--6398.

\bibitem{graham2017submanifold}
B.~Graham and L.~van~der Maaten, ``Submanifold sparse convolutional networks,''
  \emph{arXiv preprint arXiv:1706.01307}, 2017.

\bibitem{ronneberger2015u}
O.~Ronneberger, P.~Fischer, and T.~Brox, ``U-net: Convolutional networks for
  biomedical image segmentation,'' in \emph{International Conference on Medical
  image computing and computer-assisted intervention}.\hskip 1em plus 0.5em
  minus 0.4em\relax Springer, 2015, pp. 234--241.

\bibitem{gramkow2001averaging}
C.~Gramkow, ``On averaging rotations,'' \emph{Journal of Mathematical Imaging
  and Vision}, vol.~15, no. 1-2, pp. 7--16, 2001.

\bibitem{kendall2017geometric}
A.~Kendall and R.~Cipolla, ``Geometric loss functions for camera pose
  regression with deep learning,'' in \emph{Proceedings of the IEEE Conference
  on Computer Vision and Pattern Recognition}, 2017, pp. 5974--5983.

\bibitem{barfoot2019state}
T.~D. Barfoot, ``State estimation for robotics,'' 2019.

\bibitem{stoyanov2012fast}
T.~Stoyanov, M.~Magnusson, H.~Andreasson, and A.~J. Lilienthal, ``Fast and
  accurate scan registration through minimization of the distance between
  compact 3{D} ndt representations,'' \emph{The International Journal of
  Robotics Research}, vol.~31, no.~12, pp. 1377--1393, 2012.

\bibitem{li2018undeepvo}
R.~Li, S.~Wang, Z.~Long, and D.~Gu, ``Undeepvo: Monocular visual odometry
  through unsupervised deep learning,'' in \emph{2018 IEEE int. conf. on
  robotics and automation (ICRA)}, 2018, pp. 7286--7291.

\bibitem{chen1992object}
Y.~Chen and G.~Medioni, ``Object modelling by registration of multiple range
  images,'' \emph{Image and vision computing}, vol.~10, no.~3, pp. 145--155,
  1992.

\bibitem{behley2019iccv}
J.~Behley, M.~Garbade, A.~Milioto, J.~Quenzel, S.~Behnke, C.~Stachniss, and
  J.~Gall, ``{SemanticKITTI: A Dataset for Semantic Scene Understanding of
  LiDAR Sequences},'' in \emph{Proc. of the IEEE/CVF International Conf.~on
  Computer Vision (ICCV)}, 2019.

\bibitem{richter2019towards}
J.~Richter-Klug and U.~Frese, ``Towards meaningful uncertainty information for
  cnn based 6d pose estimates,'' in \emph{International Conference on Computer
  Vision Systems}.\hskip 1em plus 0.5em minus 0.4em\relax Springer, 2019, pp.
  408--422.

\bibitem{vaswani2017attention}
A.~Vaswani, N.~Shazeer, N.~Parmar, J.~Uszkoreit, L.~Jones, A.~N. Gomez,
  {\L}.~Kaiser, and I.~Polosukhin, ``Attention is all you need,'' in
  \emph{Advances in neural information processing systems}, 2017, pp.
  5998--6008.

\end{thebibliography}


\begin{thebibliography}{1}
\providecommand{\url}[1]{#1}
\csname url@rmstyle\endcsname
\providecommand{\newblock}{\relax}
\providecommand{\bibinfo}[2]{#2}
\providecommand\BIBentrySTDinterwordspacing{\spaceskip=0pt\relax}
\providecommand\BIBentryALTinterwordstretchfactor{4}
\providecommand\BIBentryALTinterwordspacing{\spaceskip=\fontdimen2\font plus
\BIBentryALTinterwordstretchfactor\fontdimen3\font minus
  \fontdimen4\font\relax}
\providecommand\BIBforeignlanguage[2]{{%
\expandafter\ifx\csname l@#1\endcsname\relax
\typeout{** WARNING: IEEEtran.bst: No hyphenation pattern has been}%
\typeout{** loaded for the language `#1'. Using the pattern for}%
\typeout{** the default language instead.}%
\else
\language=\csname l@#1\endcsname
\fi
#2}}

\bibitem{gramkow2001averaging}
C.~Gramkow, ``On averaging rotations,'' \emph{Journal of Mathematical Imaging
  and Vision}, vol.~15, no. 1-2, pp. 7--16, 2001.

\bibitem{zhang2014loam}
J.~Zhang and S.~Singh, ``Loam: Lidar odometry and mapping in real-time.'' in
  \emph{Robotics: Science and Systems}, vol.~2, no.~9, 2014.

\bibitem{kabsch1976solution}
W.~Kabsch, ``A solution for the best rotation to relate two sets of vectors,''
  \emph{Acta Crystallographica Section A: Crystal Physics, Diffraction,
  Theoretical and General Crystallography}, vol.~32, no.~5, pp. 922--923, 1976.

\end{thebibliography}

\end{document}